\documentclass[10pt,twocolumn,letterpaper]{article}

\usepackage{iccv}
\usepackage{times}
\usepackage{epsfig}
\usepackage{graphicx}
\usepackage{amsmath}
\usepackage{amssymb}

\usepackage{booktabs}
\usepackage[accsupp]{axessibility} % Improves PDF readability for those with disabilities.
\usepackage[ruled,vlined,linesnumbered]{algorithm2e}
\usepackage{comment}
\usepackage{bbm}
\usepackage{autobreak}
\usepackage{adjustbox}
\usepackage{tabularx}
\usepackage{wrapfig}
\allowdisplaybreaks

\usepackage{multirow}
\usepackage{colortbl}
\DeclareMathOperator*{\argmax}{arg\,max}

\SetCommentSty{mycommfont}

\usepackage[breaklinks=true,bookmarks=false]{hyperref}
\hypersetup{
    colorlinks=true,    % false: boxed links; true: colored links
    citecolor=green, %OrangeRed, % green,    % color of links to bibliography
    filecolor=magenta,  % color of file links
    urlcolor=red       % color of external links
}
\newcommand{\zcComment}[1]{\textcolor{black}{#1}}

\iccvfinalcopy % *** Uncomment this line for the final submission

\def\iccvPaperID{6539} % *** Enter the ICCV Paper ID here

\ificcvfinal\fi

\begin{document}

%%%%%%%%% TITLE
\title{GeT: Generative Target Structure Debiasing for Domain Adaptation}

\author{Can Zhang \qquad Gim Hee Lee\\
Department of Computer Science, National University of Singapore\\
{\tt\small can.zhang@u.nus.edu \qquad gimhee.lee@nus.edu.sg}
% \and
% Gim Hee Lee\\
% Department of Computer Science, National University of Singapore\\
% % \\
% % First line of institution2 address\\
% {\tt\small gimhee.lee@nus.edu.sg}
}

\maketitle
\ificcvfinal\thispagestyle{empty}\fi

%%%%%%%%% ABSTRACT
\begin{abstract}
    Domain adaptation (DA) aims to transfer knowledge from a fully labeled source to a scarcely labeled or totally unlabeled target under domain shift. Recently, semi-supervised learning-based (SSL) techniques that leverage pseudo labeling have been increasingly used in DA. Despite the competitive performance, these pseudo labeling methods rely heavily on the source domain to generate pseudo labels for the target domain and therefore still suffer considerably from source data bias. Moreover, class distribution bias in the target domain is also often ignored in the pseudo label generation and thus leading to further deterioration of performance. In this paper, we propose GeT that learns a non-bias target embedding distribution with high quality pseudo labels. Specifically, we formulate an online target generative classifier to induce the target distribution into distinctive Gaussian components weighted by their class priors to mitigate source data bias and enhance target class discriminability. We further propose a structure similarity regularization framework to alleviate target class distribution bias and further improve target class discriminability. Experimental results show that our proposed GeT is effective and achieves consistent improvements under various DA settings with and without class distribution bias. Our code is available at: \href{https://lulusindazc.github.io/getproject/}{https://lulusindazc.github.io/getproject/}.
    %Our code will be open-source upon paper acceptance.
\end{abstract}

%%%%%%%%% BODY TEXT

\section{Introduction}

\begin{figure}[t]
\includegraphics[scale=0.35]{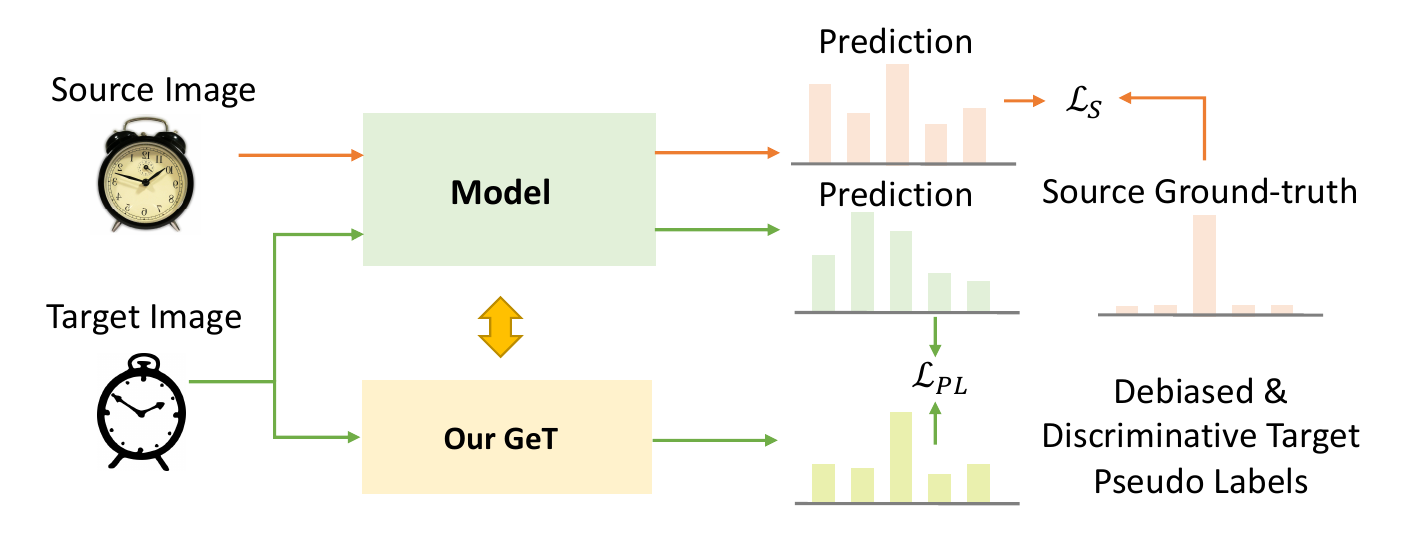}
\caption{
Our GeT is designed to generate source domain and class distribution debiased and discriminative pseudo labels for the target domain in various domain adaptation tasks.
} 
\label{fig: teaser}
\end{figure}

Despite the remarkable advances of deep learning in the last decade~\cite{resnet50, vgg, densenet, Alexnet}, the success of most deep learning-based works is based on the assumption that the data distributions of the train and test sets are similar, i.e. no domain shift. However, it is difficult to ensure no domain shift in the data distributions for many practical real-world scenarios. Consequently, many domain adaptation (DA) works~\cite{chen2018domain, kang2020pixel, sahoo2021contrast, pinheiro2019domain} have been proposed to alleviate the domain shift problem. Unsupervised DA (UDA) is the most commonly studied DA setting, where the goal is to transfer knowledge from the labeled source data to the unlabeled target data with domain shift. Other more challenging variants of UDA include partial-set DA (PDA)~\cite{cao2019learning, xu2019larger, chen2020selective} where the target label space
is a subset of the source label space, semi-supervised DA (SSDA)~\cite{saito2019semi, jiang2020bidirectional, li2020online, kim2020attract} which assumes partial target data are labeled, etc.

Most DA approaches are often either based on learning domain-invariant feature representations or directly adopting SSL techniques for knowledge transfer. It is shown in~\cite{30Therory,46theory,47theory} that the target error is bounded by the source error and the divergence between marginal distributions in the source and target domains. 
Inspired by the theoretical analysis, many works~\cite{DAN, 23ADDA, 16MCD, 12RevGrad, tang2020unsupervised} propose to learn domain-invariant feature representations using a shared feature extractor to align the source and target domains. Nonetheless, feature alignment-based methods usually suffer from the potential risk of damaging intrinsic target data discrimination. On the other hand, some recent works~\cite{chen2019domain, zhang2020label, rukhovich2019mixmatch} investigate the application of SSL techniques, e.g. MixMatch~\cite{berthelot2019mixmatch} in~\cite{rukhovich2019mixmatch}, Label Propagation~\cite{zhou2003learning} in~\cite{zhang2020label}, etc to strengthen discriminability on the unlabeled target domain. 
Although SSL-based DA methods can achieve competitive performance, they often suffer from source domain bias due to over reliance on the source domain for pseudo label generation. A recent work~\cite{liang2021domain} proposes to deal with data bias using an auxiliary target domain-oriented classifier (ATDOC) based on pseudo labeling.
It is shown that the proposed SSL regularization can work quite well in most DA scenarios.

In addition to source data bias, many DA approaches (including ATDOC) suffer significant performance drop due to class distribution bias in the target domain. Several methods~\cite{wu2019domain, tan2020class, jiang2020implicit, tachet2020domain} are proposed to alleviate class distribution bias with class-conditioned sampling~\cite{jiang2020implicit}, class-balanced self-training~\cite{tan2020class}, etc. However, as discussed in~\cite{chen2019transferability, zhao2019learning}, these methods rely on domain-invariant representation learning that can hurt intrinsic data discrimination in the target domain. Consequently, a naive adoption of these methods on SSL-based DA can lead to unreliable pseudo labels that greatly degrade performance. 
%Similar to SSL-based UDA, another critical issue of these methods %comes from is the unreliable pseudo labels and propagated errors, which %will  can greatly degrade the model performance. 
Furthermore, many existing DA methods are often designed to be task-specific and may not be %robust 
versatile enough to handle complex variants of the DA problem, e.g. PDA and SSDA.

As illustrated in Fig.~\ref{fig: teaser}, we propose GeT to generate debiased and discriminative pseudo labels to train the network on the DA tasks in this paper. Our GeT consists of an online target generative classifier and a structure similarity regularization.
%\zcComment{(while enhancing model robustness to the biased pseudo labels.)}
1) Our \textit{online target generative classifier} is a Gaussian mixture model (GMM). The class priors (i.e. mixture coefficients) and the means of the Gaussian components are the target features class distribution and prototypes, respectively. Intuitively, our generative classifier induces the target feature distribution into distinctive Gaussian components weighted by their respective class priors and thus alleviating source data bias and enhancing target class discriminability. 
We introduce a memory bank that resembles a replay buffer to efficiently store and update the class priors and feature prototypes for the classifier online in each mini-batch. 2) Our \textit{structure similarity regularization} alleviates target class distribution bias and further improves target class discriminability. To this end, we introduce an auxiliary distribution  implicitly constrained with entropy maximization to encourage balanced and discriminative pseudo labels. 
The final pseudo labels are obtained as a mixup of the pseudo labels generated by the target oriented generative classifier and the auxiliary distribution.
We jointly optimize the auxiliary distribution, the pseudo labels and the network parameters in an iterative classification expectation maximization scheme. 

We summarize our contributions as follows: 1) 
An online target oriented generative classifier is proposed to induce the distribution of the target features into distinctive Gaussian components weighted by the class priors to avoid class distribution and source data biases
while enhancing class discriminability.
2) We introduce a
structure similarity regularization that leverages an auxiliary distribution implicitly constrained with
entropy maximization to avoid the severely biased model predictions.
3) A classification expectation maximization framework is designed to jointly optimize the generative classifier with the structure similarity regularization for pseudo labels generation and train the network with the generated pseudo labels.  4) Competitive results are achieved in various DA settings on several standard benchmark datasets. 
% \vspace{-3mm}
\section{Related Work}
% \vspace{-3mm}
\paragraph{Domain Adaptation.}
Many recent deep DA works~\cite{12RevGrad,DAN,23ADDA,28CondDist2} have been proposed based on \textit{domain-invariant representation learning} using a shared feature extractor. %Among them, 
\textit{Marginal distribution alignment}~\cite{DAN,23ADDA,16MCD} and \textit{class conditional distribution alignment}~\cite{CondDist1,15CDAN,jiang2020implicit} are two representative methods which minimize various divergence measures, e.g. $\mathcal{H}$-divergence~\cite{30Therory} and maximum mean discrepancy (MMD)~\cite{gretton2006kernel} to achieve invariance. Over the years, in contrast to the widely studied covariant shift, label shift assumption is proposed in some works~\cite{chan2005word,azizzadenesheli2018regularized, lipton2018detecting, zhang2013domain} from many views, e.g. setting a prior for the label distribution~\cite{saenko2010adapting}, learning marginal label distribution with the EM algorithm~\cite{chan2005word} and designing causal/non-causal models~\cite{scholkopf2012causal, zhang2013domain, azizzadenesheli2018regularized, lipton2018detecting}. Recent works~\cite{wu2019domain, tan2020class, jiang2020implicit, tachet2020domain} exploit pseudo labels to improve the performance of DA models under class imbalance, but they still rely on learning domain-invariant representations. An auxiliary target classifier is proposed in~\cite{liang2021domain} to solve the problems of highly unreliable pseudo labels and propagated errors, but it does not consider the more practical label shift problem. Motivated by the simplicity of their framework, we aim to learn an online target-oriented generative classifier to utilize the global target data structure for improving the quality of pseudo labels under both source domain and class distribution biases. 
% Note that our method needs the source data to guide the network optimization along the training process. Although our proposed debiasing strategy mainly leverages the target data structure, the parameters of the target-oriented generative classifier %, \eg means and variances 
% are updated online together with the network parameters.
In contrast to source-free DA~\cite{kundu2020universal,lee2022confidence} that mostly freeze the source classifier during adaptation to preserve class information, our method uses a more robust pseudo-labeling strategy with the in-training source classifier optimized with data from both source and target domains. 

% \vspace{-3mm}
\paragraph{Semi-supervised Learning with Regularization.}
To leverage useful information from the unlabeled data, deep SSL introduces regularization as an auxiliary learning objective. 
Pseudo labeling~\cite{lee2013pseudo}, also known as self-training, serves as a simple and effective SSL baseline by generating pseudo labels for the unlabeled samples. A line of SSL works~\cite{tarvainen2017mean, miyato2018virtual, zhu2002learning} propose different designs of the regularization. Early SSL methods~\cite{chapelle2009semi} mainly involve Laplacian regularization, large margin regularization, etc. Recently, consistency regularization which enforces consistency between model predictions under different disturbances is becoming increasingly popular. Another widely used regularization strategy is minimum entropy~\cite{grandvalet2004semi} that aims to push model predictions to be sharp and prevent predicted label distribution from being too balanced. Moreover, regularization is also studied in 
the recent works on DA~\cite{cui2020towards, jin2020minimum, chen2019domain}. It is a special case of transductive SSL for efficiently improving domain alignment performance without explicitly designing DA strategies. In contrast, we study soft pseudo label regularization by learning target data distributions from the layer-wise feature representations without any modification on the architecture nor applying perturbations on the data or model parameters. 
% \vspace{-3mm}
\paragraph{Generative Classifier.}
It has been investigated in some works that inducing generative classifiers on the pretrained deep model for various tasks, e.g. speech recognition in~\cite{hermansky2000tandem}, novelty detection in~\cite{lee2018simple} and learning with noisy label in~\cite{lee2019robust}. A previous UDA work~\cite{lu2020stochastic} studies the stochastic classifier to improve the generalization ability of Maximum Classifier Discrepancy (MCD)~\cite{16MCD} for feature alignment. Another related work~\cite{tanwisuth2021prototype} proposes a bi-directional prototype-oriented conditional transport approach to align the target features to the source prototypes. In contrast to these methods that focus on aligning feature distributions in two domains, we introduce the generative classifier as a regularization approach to improve the quality of pseudo labels and enhance the model robustness to the class imbalance. 
\section{Problem Formulation}
%\label{Method:problem}
\paragraph{Definitions.} \textbf{DA} aims to deal with the domain shift between a set of labeled source data $\mathcal{D}_\mathcal{S}=\{(x_i^s,y_i^s)\}_{i=1}^{N_s}$ and a set of target data $\mathcal{D}_\mathcal{T}=\mathcal{D}_\mathcal{T}^u\cup \mathcal{D}_\mathcal{T}^l$, where $|\mathcal{Y}_S|=C$ and $y_i^s \in \{1, \dots, C\}$ represent the source label space with $C$ classes, and $\mathcal{D}_\mathcal{T}^u$ and $\mathcal{D}_\mathcal{T}^l$ denote the unlabeled and labeled target data, respectively. \textbf{UDA} and \textbf{PDA} assume empty target labeled set $\mathcal{D}_\mathcal{T}^l = \emptyset$ and unlabeled target set $\mathcal{D}_\mathcal{T}^u=\{x_i^t\}_{i=1}^{N_t^u}$. \textbf{SSDA} assumes partial target data are labeled, denoted by $\mathcal{D}_\mathcal{T}^l=\{x_i^t,y_i^t\}_{i=1}^{N_t^l}$.
Furthermore, UDA and SSDA assume a common target and source label space, i.e. $\mathcal{Y}_U=\mathcal{Y}_S$, and PDA assumes that the target label space is a subset of the source label space, i.e. $\mathcal{Y}_U \subset \mathcal{Y}_S$. The objective is to predict the labels $\{\hat{y}^t\}$ of the unlabeled target samples $\{x^t\} \in \mathcal{D}_\mathcal{T}^u$ by utilizing the labeled source data $\mathcal{D}_\mathcal{S}$ and limited target labeled data $\mathcal{D}_\mathcal{T}^l$ if available. Based on the assumption from~\cite{10JAN,21DC}, 
there exists a shared feature space across domains. 

% \vspace{-3mm}
\paragraph{Objective.} 
Our goal is to learn a network $g(\phi(x; \theta_f); \theta_g)$ that is able to handle the source data for the DA tasks with the SSL pseudo labeling approach under class distribution bias. 
%\zcComment{(Our goal is to reduce data bias inherent in the DA tasks designed with the pseudo labeling approach by decoupling the target data structure from the shared space, and enhance the model $g(\phi(x; \theta_f); \theta_g)$ robustness to the target class distribution bias.)}
$\phi(x; \theta_f): x \mapsto f$ denotes the feature embedding function that maps $x$ to the shared feature space $f$. $g(f; \theta_g): f \mapsto \mathcal{Y}$ denotes the classifier that maps features $f$ to the label space $\mathcal{Y}$.
\section{Our Method}

\begin{figure*}[ht]
% \vspace{-2mm}
\centering
\makebox[0.9\textwidth][c]{\includegraphics[scale=0.5]{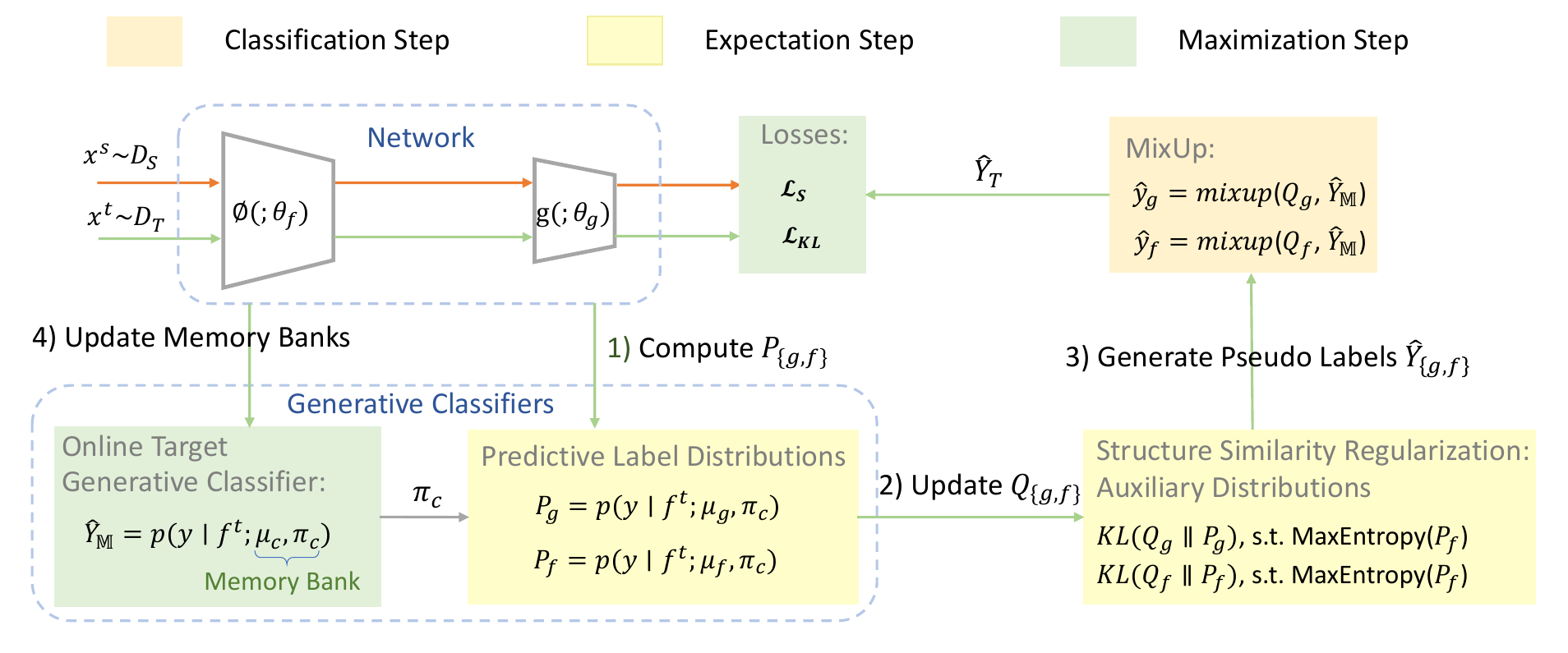}}
\caption{%Overview of the training procedure for GeT. 
\textbf{Overview of GeT.} Our GeT consists of a \textit{online target generative classifier} and a \textit{structure similarity regularization} to generate debiased and discriminative pseudo labels for the supervision of the network on various DA tasks.
} 
% \vspace{-5mm}
\label{fig: framework}
\end{figure*}

\paragraph{\textit{Overview}.} 
% We propose a regularization approach for improving the quality of pseudo labels under source data and class distributions biases. 
\zcComment{Our GeT is designed to generate debiased pseudo labels while improving model robustness towards %noises in
noisy pseudo labels.
As illustrated in Fig.~\ref{fig: framework}, our GeT iteratively optimizes the pseudo label generation and trains the network using Classification Expectation Maximization (CEM). 
In the maximization step, we train the network with the generated pseudo labels, and updates the memory bank with the updated network. 
Specifically, we proposes a generative classifier to generate pseudo-labels $\hat{Y}_{\mathbbm{M}}$ for alleviating data bias from the source domain and class bias in the target domain. The parameters of our generative classifier are efficiently updated by constructing two memory banks: 1) a feature prototypes $\{\mu_c\}_{c=1}^C$, and 2) a class prior distributions $\{\pi_c\}_{c=1}^C$. %On top of 
In addition to modeling the target data structure for improving the quality of pseudo labels, we further introduce the structure similarity regularization for improving model robustness towards noises. 
In the expectation step, 
% we compute the predictive label distributions of the model and introduce the structure similarity regularization. 
we compute the predictive label distributions $P_f$ and $P_g$ with GMMs in the feature space $\mathcal{N}(x^t|\mu_f)$ and the output space $\mathcal{N}(x^t|\mu_g)$. The corresponding auxiliary target distributions $Q_f$ and $Q_g$ are defined with the empirical distribution of the samples being assigned to the clusters, and updated to enforce balanced assignments.
In the classification step, we generate the optimal pseudo labels from the generative classifier and auxiliary distributions by mixing up data structure-wise knowledge and model-wise knowledge. Using the pseudo-labels $\hat{Y}_\mathbbm{M}$ and target variables $Q_{\{f,g\}}$, we infer pseudo labels $\hat{y}_{\{f,g\}}$ for the target loss $\mathcal{L}_{KL}^{\{f,g\}}$ and additionally adopt the source loss $\mathcal{L}_{S}$ to optimize the network.}

\subsection{Generative Classifier}
\zcComment{The presence of data bias from the source domain has been shown to degrade the quality of pseudo labels for model adaptation in the target domain \cite{wu2019domain, tan2020class, jiang2020implicit, tachet2020domain}.
To alleviate data bias from the source domain, we
propose a target domain-oriented classifier that can fully exploit the target data structure to generate reliable pseudo labels for the unlabeled target data. Although the prototype-based classifier~\cite{liang2021domain} helps to alleviate the data bias from the source domain, it is sensitive to the class distributions in the target domain by favoring dominant classes over the minority classes. Consequently, the class-imbalance %assumption
problem further motivates us to extend the prototype-based classifier to a Gaussian mixture model that can learn the intrinsic target data distributions for balancing the predictive label distributions.}

We model the distribution of the feature embeddings randomly sampled from the target domain $f^t \sim \mathcal{D}_{\mathcal{T}}$ with a Gaussian mixture model given by:
\begin{equation}
p(f^t) = \sum_c \pi_c \mathcal{N}(f^t \mid \mathbf{\boldsymbol{\mu}}_c, \Sigma_c),
\end{equation}
where $\pi_c=p(y_c)$ is the class prior, i.e. mixing coefficient, and $\{\mathbf{\mu}_c, \Sigma_c\}$ are the class prototype, i.e. mean and covariance of the Gaussian component $\mathcal{N}(f^t \mid \mu_c, \Sigma_c)$. We define our generative classifier as the posterior probability of class $y_c$ given $f^t$:
\begin{equation}
\begin{split}
\small
     p(y_c \mid f^t) 
     &= \frac{\pi_cp(f^t \mid \boldsymbol{\mu}_c, \Sigma_c)}{\sum_{c'}\pi_{c'}p(f \mid \boldsymbol{\mu}_{c'}, \Sigma_c)} \\
     &=\sigma(\log \pi_c + \frac{s(f^t, \boldsymbol{\mu}_c)}{\tau}),
\end{split}
\label{eq:posterior}
% \vspace{-2mm}
\end{equation}
where $\sigma(\cdot)$ represents the softmax function, $s(\cdot, \cdot)$ measures the similarity between feature embeddings and prototypes (cosine similarity is adopted by default), and $\tau$ is a temperature hyperparameter analogous to the class covariance $\Sigma_c$. $\frac{s(\cdot, \cdot)}{\tau}$ gives the log-likelihood $\log p(f^t \mid \boldsymbol{\mu}_c, \Sigma_c)$. 

\subsection{Online Target Generative Classifier}
We maintain an online target generative classifier using the features from the target domain, where the parameters, \ie the class prior $\pi_c$ and the class prototypes $\boldsymbol{\mu}_c$ over the entire distribution of the target embeddings are updated efficiently online using a memory bank.
\zcComment{In contrast, many early works~\cite{liang2020we, yue2021prototypical} naively compute the feature cluster centroids based on unsupervised feature clustering, \eg k-means, which requires the computationally expensive extraction of all the feature embeddings with the feature extractor for clustering. 
% To reduce computations, they choose to re-compute the clusters only at the beginning of each epoch.
}
%Specifically, we use a memory bank to mitigate the computational complexity for computing the class prior $\pi_c$ and the class prototypes $\boldsymbol{\mu}_c$ over the entire distribution of the target embeddings.

\paragraph{Class priors $\pi_c$.}
Given the $i^\text{th}$ batch of target data $\{f^t_{j,i}\}_{j=1}^{\mathcal{B}^t}$, we first generate the pseudo label $\hat{y}^t_{\mathbbm{M},j}=\argmax_c p(y \mid f^t_{j,i};~ \boldsymbol{\mu_{1:C}}, \pi_{1:C}) $ of each data $f^t_{j,i}$ with the current class prototypes $\boldsymbol{\mu}_{1:C}$ and class prior $\pi_{1:C}$ in the memory bank. We then use $\hat{y}^t_{\mathbbm{M},j}$ in a mixup strategy to generate more reliable pseudo labels to train the network $g(\phi(x; \theta_f); \theta_g)$ (\cf Section \ref{sec:optimization} for more details). The class prior is updated as:
\begin{equation}
%\small
\begin{split}
    & \pi_{c} \leftarrow (1-\gamma_{\pi})\pi_{c} + \gamma_{\pi} \bar{P}, \\
    % \quad 
    & \text{where} \quad \bar{P}  = \frac{1}{\mathcal{B}^t}\sum_{j=1}^{\mathcal{B}^t}\frac{s(f^t_{j,i}, \boldsymbol{\mu}_c^g)}{\tau}, %p(y \mid f^t_{j,i}; \boldsymbol{\mu}^{g}_{1:C}, \pi_{1:C}).
    \end{split} 
\label{eq:prior_deriv}
\end{equation}
$\boldsymbol{\mu}_c^{g} = \theta_{g_c} \backslash b_c$ is the weights of the linear classifier $g(.; \theta_g)$ corresponding the class $c$ without the bias terms $b_c$, and $\gamma_{\pi}$ is the memory decay coefficient. We refer to $\boldsymbol{\mu}_c^{g}$ as the classifier prototype.  
The intuition behind this formulation is that the label prior for each class can be estimated by averaging the likelihoods over all target features and the memory bank eases on the computational complexity. 
Note that we initialize the class prior as a uniform distribution to $\pi_{c}=\frac{1}{C}$.

\paragraph{Class prototypes $\boldsymbol{\mu}_c$.}
% \zcComment{
% To compute the class prototypes, early works~\cite{liang2020we, yue2021prototypical} naively compute the feature cluster centroids based on all feature embeddings according to the class labels. This method requires to extract all the feature embeddings with the feature extractor first, which has the heavy computation cost.}
Different from previous works which compute the prototypes based on all features according to the class labels,
we propose to construct the class prototypes $\boldsymbol{\mu}_{1:C}$ on the entire target feature space online using a memory bank. We derived the update of the learnable class prototypes $\boldsymbol{\mu}_{1:C}$ from the conditional distribution $p(\boldsymbol{\mu}_{c} \mid  f^t_{1:i})$ based on the entire historical inputs $\{f^t_{1:i}, y_c\}$ as follows:
\begin{equation}
% \small
    \begin{split}
        & \log p(\boldsymbol{\mu}_{c} \mid  f^t_{1:i}) 
         \propto \log p(\boldsymbol{\mu}_{c} \mid  f^t_{1:i-1}) + \log p(f^t_{i} \mid  \boldsymbol{\mu}_{c}) := \\  
        %  \approx P(f^t \mid  \boldsymbol{\mu}_{c,i}, \Sigma_i ) P(\boldsymbol{\mu}_{c,i} \mid \boldsymbol{\mu}_{c,i-1}^u, \sigma^2_{pt}+\Sigma_{i-1}^u),
          & \boldsymbol{\mu}_{c} 
           \leftarrow (1-\gamma_{\boldsymbol{\mu}}) * \boldsymbol{\mu}_{c} + \gamma_{\boldsymbol{\mu}} *\bar{\boldsymbol{\mu}}_{c},~\text{where}~\bar{\boldsymbol{\mu}}_{c}= \frac{1}{\mathcal{B}^t_c}\sum_{j=1}^{\mathcal{B}^t}\mathbbm{1}_{c,j}f_{j,i}^t.
    \end{split}
\label{eq:mu_deriv}
\end{equation}
$\mathbbm{1}_{c,j} = \mathbbm{1}[p(y_c \mid f^t_{j,i}) \geq p(y_{c'} \mid f^t_{j,i}), \forall c' \in C]$ selects the features that give the highest class probability for class $c$ to update the class prototype $\mathbf{\mu}_c$. $\gamma_{\mu}$ is the memory decay coefficient defined in the same way as $\gamma_{\pi}$.

% \vspace{-3mm}
\paragraph{Remarks:} Our generative classifier mitigates source domain bias since it is based solely on the target features. Furthermore, our Bayesian formulation of the classifier encourages discriminative features with the class prior $\pi_c$.

\subsection{Structure Similarity Regularization}
%\revision{
Although our generative classifier can mitigate source data bias and encourage discriminative features, it still risks wrong assignments of features from the scarce classes under class distribution bias. To this end, we introduce a structure similarity regularization which alternates between optimizing the KL-divergence of an auxiliary distribution to the predictive label distributions and the network parameters to encourage balanced and discriminative assignments of features into their respective classes.

\paragraph{Classifier label distribution $P_{g}$.} We estimate the label distribution %$\boldsymbol{P}_g^t \overset{\Delta}{=} \{P_{g}(y \mid f_j^t)\}_{j=1}^{N_t^u}$
$P_{g}:=p(y~\mid~f_j^t;~\boldsymbol{\mu}_{1:C}^g,~\pi_{1:C})$
from the features of the unlabeled target data $\{f^t_{j}\}_{j=1}^{N_t^u}$, 
the classifier prototypes $\boldsymbol{\mu}_{1:C}^g$ and class prior distribution $\pi_{1:C}$.
% given by the weights $\theta_g$ of the linear classifier.
%For brevity, we write $P_{g}(y = c \mid f_j^t)$ as $P_{g}(y_c \mid f_j^t)$. Motivated by~\cite{ghasedi2017deep}, we propose $\boldsymbol{Q}_g^t \overset{\Delta}{=} \{Q_{g}(y \mid f_j^t)\}_{j=1}^{N_t^u}$ as a regularized auxiliary distribution to approximate $\boldsymbol{P}_g^t$. %Our GeT model then performs the optimization by the following objective:
% The target variable $\boldsymbol{Q}_g^t$ can be arbitrary distribution. 
A naive optimization of $P_g$ with the pseudo labels $\hat{Y}_{\mathbb{M}}^t$ produced by our generative classifier can cause degenerate solutions where data from scarce classes are assigned wrongly due to class distribution bias. Motivated by~\cite{ghasedi2017deep}, we introduce an auxiliary distribution $Q_{g}$ and minimize the following loss:
\begin{equation}
\begin{split}
    % \small
     \mathcal{L}_\text{KL}^{g}=
    % \text{KL}(\boldsymbol{Q}^t ||\boldsymbol{P}^t) + \sum_{c=1}^{C} \varrho_{c}^{t} \log \varrho_{c}^{t}, 
    \frac{1}{N_t^u}\operatorname{KL}(Q_{g} \| P_{g}) + 
     \sum_{c=1}^{C} \bar{Q}_{g_c} \log \bar{Q}_{g_c}
\end{split}
\label{eq:KL_div}
\end{equation}
over $\theta_f$, $\theta_g$ and $Q_g$. The first term compute the KL-divergence between the discrete posteriors $P_g$ and $Q_g$. The second term %$\mathcal{R}_{g}$ 
plays the role of confidence penalty by encouraging entropy maximization of the label distribution in the target domain. $\bar{Q}_{g_c} =\frac{1}{N^u_t}\sum_{j=1}^{N^u_t}Q_{g}(y_{c} \mid f_{j}^t)$ is defined to %estimate
be the target class proportions. Intuitively, the unlabeled data are more likely to be assigned to the prototypes corresponding to the dominant classes or the prototypes that are much closer to the target features. 
% In contrast to~\cite{ghasedi2017deep}, we do not adopt the cluster assignments $P_{g}(y \mid f^t)$ with a uniform prior due to the class imbalance. Instead, we incorporate the discrete prior distribution $\pi^\mathbbm{M}$ into the predictive label distribution $\boldsymbol{P}_g^t $ through the lens of Bayes' rule from Eq.~\ref{eq:posterior}.
The empirical label distribution $\bar{Q}_{g}$ of the regularized auxiliary distribution is enforced to have balanced assignments by %$\mathcal{R}_{g}$
the second term, which is equivalent to using the KL-divergence between $\bar{Q}_{g}$ and a uniform prior distribution.
%In aware of no prior knowledge of the target class distribution $P(\boldsymbol{\mu})$, we thus rely on the second term to encourage the cluster size balance. 
% The first term compute the KL-divergence between discrete posteriors $\boldsymbol{P}^t$ and $\boldsymbol{Q}^t$ as:
% \begin{equation}
% \small
%     \text{KL}(\boldsymbol{Q}^t||\boldsymbol{P}^t) = \frac{1}{N_t^u}\sum_{j=1}^{N_t^u}\sum_{c=1}^{C}q_{\theta}(\boldsymbol{\mu}_c|f_j^t) \log \frac{q_{\theta}(\boldsymbol{\mu}_c|f_j^t)}{P_{\theta}(\boldsymbol{\mu}_c|f_j^t)}.
%     \label{eq:KL_div_form}
% \end{equation}
We minimize the loss $\mathcal{L}_{\text{KL}}^g$ using an alternating optimization based on the following two steps: 
% (1) pseudo label generation $\boldsymbol{Q}^t$, and (2) retraining the network with $\boldsymbol{P}^t$ and the updated $\boldsymbol{Q}^t$
% \revision{(See our supplementary for more details of the derivation.)}
% \vspace{-3mm}

\paragraph{a) Pseudo-label generation.} We fix the network parameters $\{\theta_f, \theta_g\}$ and $P_g$ to estimate the auxiliary distribution $Q_g$. The closed-form solution of $Q_{g}$ can be derived by setting the gradient of the optimization objective from Eq.~\ref{eq:KL_div} as zero, \ie:
\begin{equation}
% \small
    Q_{g}(y_{c} \mid f_{j}^t) = \frac{P_{g}(y_{c} \mid f_{j}^t) / (\sum_{j=1}^{N^u_t}P_{g}(y_{c} \mid f_{j}^t))^{\frac{1}{2}}}{\sum_{c'=1}^{C}P_{g}(y_{c'} \mid f_{j}^t) / (\sum_{j=1}^{N^u_t}P_{g}(y_{c'} \mid f_{j}^t))^{\frac{1}{2}}}.
    \label{eq:auxiliary_distri}
\end{equation}

% \vspace{-3mm}
\paragraph{b) Network retraining.} By fixing $Q_g$, the second term in Eq.~\ref{eq:KL_div} reduces to a constant value and thus giving rise to a cross entropy loss using $Q_g$ as the soft label for network optimization:
\begin{equation}
% \small
    \min_{\theta_f, \theta_g} 
    -\frac{1}{N_t^u}\sum_{j=1}^{N_t^u}\sum_{c=1}^{C}Q_{g}(y_c \mid f_j^t) \log P_{g}(y_c \mid f_j^t).
    \label{eq:KL_div_train}
\end{equation}
Note that the pseudo-label generation step will be included in the C-step and the network retraining step will be included in the M-step of the final CEM optimization shown in the next section. 

% \vspace{-3mm}
\paragraph{Embedding label distribution $P_{f}$.} Based on the clustering assumption in the feature space, we also introduce a set of learnable embedding prototypes $\boldsymbol{\mu}_{1:C}^f$ to discover the target feature discrimination. %Assume $\{\boldsymbol{\mu}_c^v\}_{c=1}^{C}$ be the learnable prototypes, 
% We approximate the learnable embedding prototypes with a posterior 
% %$\widetilde{P}^t(\boldsymbol{\mu}^v \mid f_t)$
% $\boldsymbol{\mu}_c^f \approx \widetilde{P}_{\theta}(y_c \mid f^t)$ as the soft label assignments for a target feature $f^t_j$ according to the similarity between the single instance and embedding prototypes in the embedding space. The probability of $f^t_j$ being assigned to the $c^\text{th}$ prototype is defined as:
We compute the label distribution $P_{f} := p(y~\mid~f_j^t;~\boldsymbol{\mu}_{1:C}^f,~\pi_{1:C})$ %for a target feature $f^t_j$ according to the similarity between the single instance and embedding prototypes in the embedding space as Eq.~\ref{eq:posterior}.
% \begin{equation}
% \small
%     {P}_{f}(y_c \mid f_t) =\exp{(\frac{s(f^t_j,\boldsymbol{\mu}^f_c)}{\eta})} / \sum_{c'=1}^{C}\exp{(\frac{s(f^t_j,\boldsymbol{\mu}^f_{c'})}{\eta})}.
%     \label{eq:posterior_v}
% \end{equation}
%We now define ${\boldsymbol{P}}_f^t \overset{\Delta}{=} \{{P}_f(y \mid f^t_j)\}_{j=1}^{N^u_t}$, 
and introduce an auxiliary distribution ${Q}_f$. %The objective of feature discrimination based on ${\boldsymbol{P}}_f^t$ and ${\boldsymbol{Q}}_f^t$ is defined as:
We then minimize the following loss:
\begin{equation}
\begin{split}
    % \small
    \mathcal{L}_\text{KL}^{f}=
    % \text{KL}(\widetilde{\boldsymbol{Q}}^t||\widetilde{\boldsymbol{P}}^t) + \sum_{c=1}^{C} \tilde{\varrho_{c}}^{t} \log \tilde{\varrho}_{c}^{t},
     \frac{1}{N_t^u}\operatorname{KL}(Q_f \| P_f) + \sum_{c=1}^{C} \bar{Q}_{f_c} \log \bar{Q}_{f_c}.
\end{split}
\label{eq:KL_div_f}
\end{equation}
over $\theta_f$, $\boldsymbol{\mu}_{1:C}^{f}$, and ${Q}_{f}$. The regularization term is defined as $\bar{Q}_{f_c} = \frac{1}{N^u_t}\sum_{j=1}^{N^u_t}{Q}_{f}(y_c \mid f_{j}^t)$. 
% We optimize $\mathcal{L}^f_{\text{KL}}$ over $Q_f$ and then $\theta_f$ and $\boldsymbol{\mu}_{1:C}^f$ iteratively. 
$\boldsymbol{\mu}_{1:C}^f$ are re-initialized at each epoch using the class prototypes $\boldsymbol{\mu}_{1:C}$ from the memory bank. We apply the same alternating optimization strategy for $\mathcal{L}_{\text{KL}}^g$ on $\mathcal{L}_{\text{KL}}^f$, where the auxiliary distribution $Q_f$ is used as the soft label for optimizing the feature embedding network parameters $\theta_f$ and embedding prototypes $\boldsymbol{\mu}_{1:C}^f$.

\subsection{Optimization} \label{sec:optimization}
Given the labeled samples $\{x_j^s,y_j^s\}$ from the source domain $\{X_S,Y_S\}$ and unlabeled samples $\{x_j^t\}$ from the target domain $X_T$, 
% our GeT can be written as the following regularized classification maximum likelihood (CML):
% \begin{equation}
%     \small
%     \begin{split}
%         % \max_{\theta_f,\theta_g, \hat{Y}_T} \log\tilde{\mathcal{L}}_{C}-\mathcal{R}_{C}
%         \max_{\theta_f, \theta_g, \hat{Y}_T} 
%         % \frac{1}{N_s} \sum_{j=1}^{N_s}\sum_{c=1}^{C}y^{s, c}_{j} \log p_{\theta}(y_j^s=c \mid x_j^s) +\frac{1}{N_t^u}\sum_{j=1}^{N_t^u}\sum_{c=1}^{C} \hat{y}^{t, c}_j \log p_{\theta}(y_j^t=c \mid x_j^t)
%         \log{\mathcal{L}}^{s}(X_S,Y_S)+\log{\mathcal{L}}^{t}(X_T,\hat{Y}_T)+\mathcal{R},
%     \end{split}
%     \label{eq: overall_loss}
% \end{equation}
% where ${\mathcal{L}}^{s}$ is the source likelihood, ${\mathcal{L}}^{t}$ is the target likelihood and $\mathcal{R} = -\mathcal{R}_g-\mathcal{R}_f$ is the confidence regularizer. The regularized classification maximum likelihood of 
our GeT model is alternatively optimized by the CEM steps: 

% \vspace{-3mm} 
\paragraph{\textit{E-Step}:}
Compute the posterior probabilities classifier $P_{g}:=p(y~\mid~f_j^t;~\boldsymbol{\mu}_{1:C}^g,~\pi_{1:C})$ and embedding $P_{f}:=p(y~\mid~f_j^t;~\boldsymbol{\mu}_{1:C}^f,~\pi_{1:C})$ label distributions from the current batch of target features $\{f^t_{j,i}\}_{j=1}^{\mathcal{B}^t}$, class prior $\pi_{1:C}$, and classifier $\boldsymbol{\mu}_{1:C}^g$ and embedding $\boldsymbol{\mu}_{1:C}^f$ prototypes.

% \vspace{-1mm}
\paragraph{\textit{C-Step}:}
Fixing the network parameters $\{\theta_f, \theta_g\}$, we solve the pseudo label generation step on the objectives from Eq.~\ref{eq:KL_div} and Eq.~\ref{eq:KL_div_f} to get the auxiliary distributions $Q_{\{g,f\}}$. We then fully utilize data structure knowledge from both domains to get the final pseudo labels $\hat{Y}_T=\{\hat{y}_g^t, \hat{y}_f^t\}$ by applying mixup on the soft labels from $Q_{\{f,g\}}$ and the pseudo-labels $\hat{Y}_{\mathbbm{M}}^t$ from our generative classifier, i.e.:
%pseudo labels $\hat{Y}_{\{g,f\}}^t$, which is simplified to the classification maximum likelihood (CML) objective as follows:
\begin{equation}
\small
    \hat{y}_{\{g,f\},j}^{t,c}=(1-\gamma_Q)Q_{\{g,f\}}(y_c \mid f_{j}^t) + \gamma_Q \hat{y}^{t,c}_{\mathbbm{M},j},
    \label{eq:infer_label}
\end{equation}
where $\gamma_Q$ is the coefficient for the mixup. 

% \vspace{-1mm} 
\paragraph{\textit{M-Step}:}
Fixing $\hat{Y}_T=\{\hat{y}_g^t, \hat{y}_f^t\}$, we use the gradient ascent to update the network parameters $\theta_f$ and $\theta_g$, and the learnable embedding prototypes $\boldsymbol{\mu}_{1:C}^f$:
\begin{equation}
\small
\begin{split}
    &\max_{\theta_f,\theta_g,\boldsymbol{\mu}_{1:C}^f} \frac{1}{N_s} \sum_{j=1}^{N_s}\sum_{c=1}^{C}y_{j}^{s,c} \log p_{\theta}(y_{j}^{s,c} \mid x_j^s) + \\ & \frac{1}{N_t^u}\sum_{j=1}^{N_t^u}\sum_{c=1}^{C} \hat{y}^{t,c}_{g,j} \log P_{g}(y_c \mid f_j^t) +
        \hat{y}^{t,c}_{f,j} \log P_{f}(y_c \mid f_j^t),
\end{split}
\label{eq:cross_entropy}
\end{equation}
where $p_{\theta}(y \mid x^s)=g(\phi(x^s; \theta_f); \theta_g)$ denotes the output label predictions of source data $x^s$ from the network. The second and third terms are adapted from the network retraining step mentioned in the previous section.
\section{Experiment}

\begin{table*}[t!]
\caption{Classification accuracy (\%) on Office-Home for UDA (ResNet-50). } 
% \begin{center}
\centering
\scriptsize
% \scalebox{0.5}{
% \begin{tabular*}{0.95\columnwidth}{@{\extracolsep{\stretch{1}}}*{1}{l|cccccccccccc|c}@{}}
\begin{adjustbox}{width=1\textwidth}
\begin{tabular}{l|cccccccccccc|c }
\hline
Method & A$\to$C & A$\to$P  & A$\to$R  & C$\to$A  & C$\to$P  & C$\to$R  & P$\to$A  & P$\to$C  & P$\to$R & R$\to$A  & R$\to$C  & R$\to$P  & Mean{\cellcolor[rgb]{0.902,0.902,0.902}}\\
\hline
ResNet-50 &44.9 &66.3 &74.3  &51.8 &61.9 &63.6 &52.4 &39.1 &71.2 &63.8 &45.9 &77.2 &59.4 {\cellcolor[rgb]{0.902,0.902,0.902}} \\
% ResNet-50 &44.9&66.3&74.3&51.8&61.9&63.6&52.4&39.1&71.2&63.8&45.9&77.2&59.4{\cellcolor[rgb]{0.902,0.902,0.902}} \\
MinEnt~\cite{grandvalet2004semi}&51.0  &71.9  & 77.1 & 61.2 &69.1  &70.1 &59.3 &48.7  & 77.0 & 70.4 &53.0  &81.0  &65.8 {\cellcolor[rgb]{0.902,0.902,0.902}} \\
% MinEnt~\cite{grandvalet2004semi}&51.0&71.9&77.1&61.2&69.1&70.1&59.3&48.7&77.0&70.4&53.0&81.0&65.8{\cellcolor[rgb]{0.902,0.902,0.902}} \\
BNM~\cite{cui2020towards}&56.7 &77.5 &81.0  &67.3  &76.3  &77.1 &65.3  & 55.1 &82.0  &73.6  &57.0  &84.3  &71.1 {\cellcolor[rgb]{0.902,0.902,0.902}} \\
MCC~\cite{jin2020minimum}& 56.3 & 77.3 &80.3 &67.0 &77.1 &77.0 &66.2 &55.1 &81.2 &73.5 &57.4 &84.1 &71.0  {\cellcolor[rgb]{0.902,0.902,0.902}}\\
PL &54.1 &74.1 &78.4 &63.3 &72.8 &74.0 &61.7 &51.0 &78.9 &71.9 &56.6  &81.9  &68.2 {\cellcolor[rgb]{0.902,0.902,0.902}} \\
ATDOC-NC~\cite{liang2021domain}&54.4 &77.6 &80.8 &66.5 &75.6 &75.8 &65.9 &51.9 &81.1 &72.7 &57.0 &83.5 &70.2  {\cellcolor[rgb]{0.902,0.902,0.902}}\\
GeT &59.4 &\textbf{79.6} &\textbf{82.9} &\textbf{71.4} &\textbf{79.8}  &\textbf{79.8} &69.7 &56.2 &83.5  &73.9 &60.1 &86.0 & 73.5  {\cellcolor[rgb]{0.902,0.902,0.902}} \\
\hline
CDAN+E~\cite{long2018conditional}&54.6 &74.1 &78.1 &63.0 &72.2 &74.1 & 61.6 &52.3 &79.1  &72.3  &57.3  & 82.8 &68.5  {\cellcolor[rgb]{0.902,0.902,0.902}}\\
+ BNM~\cite{cui2020towards}&58.1 &77.2 &81.1 &67.5 &75.3  &77.2  &65.5  &56.8 &82.6  & 74.1 &59.9  &84.6  &71.7  {\cellcolor[rgb]{0.902,0.902,0.902}}\\
+ MCC~\cite{jin2020minimum}&58.9  &77.6 &80.7 &67.0 &75.1 &77.1 &65.8 &56.8 &82.2 &73.9 &59.8 &84.5 &71.6 {\cellcolor[rgb]{0.902,0.902,0.902}}\\
+ PL&57.3 &76.6 &79.2 &66.6 &74.0 &76.6 &66.1 &53.6 &81.0 &74.3 &58.9 &84.2 &70.7  {\cellcolor[rgb]{0.902,0.902,0.902}}\\
+ ATDOC-NC~\cite{liang2021domain}&55.9 &76.3 &80.3 &63.8 &75.7 &76.4 &63.9 &53.7 &81.7 &71.6 &57.7 &83.3 &70.0 {\cellcolor[rgb]{0.902,0.902,0.902}} \\
+ GeT&\textbf{60.5} &78.8 &82.6 &69.1 &79.7 &78.8 &\textbf{69.5} &\textbf{59.3} &\textbf{84.6} &\textbf{75.2} &\textbf{62.3} &\textbf{88.0} & \textbf{74.0}  {\cellcolor[rgb]{0.902,0.902,0.902}} \\
\hline
SAFN~\cite{xu2019larger}&52.0&71.7&76.3&64.2&69.9&71.9&63.7&51.4&77.1&70.9&57.1&81.5&67.3 {\cellcolor[rgb]{0.902,0.902,0.902}}\\
% CADA-P~\cite{kurmi2019attending}&56.9&76.4&80.7&61.3&75.2&75.2&63.2&54.5&80.7&73.9&61.5&84.1&70.2 {\cellcolor[rgb]{0.902,0.902,0.902}}\\
% DCAN~\cite{li2020domain}&54.5&75.7&81.2&67.4&74.0&76.3&67.4&52.7&80.6&74.1&59.1&83.5&70.5 {\cellcolor[rgb]{0.902,0.902,0.902}}\\
SHOT~\cite{liang2020we}&57.1&78.1&81.5&68.0&78.2&78.1&67.4&54.9&82.2&73.3&58.8&84.3&71.8{\cellcolor[rgb]{0.902,0.902,0.902}}\\
SCDA~\cite{li2021semantic} & 57.5 & 76.9 & 80.3 & 65.7 & 74.9 & 74.7 & 65.5 & 53.6& 79.8 & 74.5 & 59.6 & 83.7& 70.5{\cellcolor[rgb]{0.902,0.902,0.902}}\\
DALN~\cite{chen2022reusing} & 57.8 & 79.9 & 82.0 & 66.3 & 76.2 & 77.2 & 66.7 & 55.5 & 81.3 & 73.5 & 60.4 & 85.3 & 71.8 {\cellcolor[rgb]{0.902,0.902,0.902}}\\
\zcComment{
FixBi~\cite{na2021fixbi}} & 58.1 & 77.3 & 80.4 & 67.7 & 79.5 & 78.1 & 65.8 & 57.9 & 81.7 & 76.4 & 62.9 & 86.7 & 72.7 {\cellcolor[rgb]{0.902,0.902,0.902}}\\
\hline
\end{tabular}
% }
% \end{center}
\end{adjustbox}
\label{tab: office-home}
% \vspace{-3mm}
\end{table*}

\subsection{Datasets and Experimental Setting}
\label{sec:ExperimentalSetup}
% \vspace{-1mm}
\paragraph{Datasets.}
\textbf{Office-31}~\cite{3dataoffice} includes three domains: Amazon (A), DSLR (D) and Webcam (W), and contains a total of 4,110 images covering 31 categories. A combination of six pairs of source-target domain settings are evaluated. \textbf{Office-Home}~\cite{5dataclef} includes 4 domains: Artistic (Ar), Clip Art (CI), Product (Pr) and Real-World (Re) with 65 categories, where there are $\sim$15,500 images in total. \textbf{VisDA-2017} \cite{4datavisda} is a challenging dataset due to the big domain shift between the synthetic images (152,397 images from VisDA) and the real images (55,388 images from COCO). 
\textbf{DomainNet-126} is constructed in~\cite{saito2019semi} by selecting 126 classes across 4 domains, i.e. Real (R), Clipart (C), Painting (P) and Sketch (S), from the largest UDA dataset DomainNet~\cite{peng2019moment}.

% \vspace{-3mm}
\paragraph{Implementation details.}
Following~\cite{liang2021domain}, 
we use ResNet-50 pretrained on the ImageNet as the backbone and utilize a mini-batch SGD with momentum 0.9 and weight decay $1e^{-3}$. The learning rate follows the schedule as $\eta_{i}=\eta_{0}(1+\omega \frac{i}{I_{max}})^{-\alpha}$, where $\omega=10$, $\alpha=0.75$, and $\eta_0$ is the initial learning rate. We set $\eta_0=0.001$ for the target-specific bottleneck layers and $\eta_0=0.01$ for the classifier. We set $\gamma_{\bar{\mu}}=0.9$ and
perform the sensitivity analysis on $\gamma_Q$ and $\gamma_{\bar{P}}$. We set the temperature hyper-parameter $\tau$ to 1 empirically. We adopt the imbalanced target class setting by following~\cite{tachet2020domain}, where only thirty percent of data from the first $[C/2]$ classes are kept to simulate class imbalance in the target domain. 
% \revision{It should be noted that all baselines are reproduced in the same class-imbalanced setting as our GeT to make fair comparisons, and thus different from their originally reported results evaluated under the strict assumption of the balanced data distribution}. 

\begin{table}[t!]
% \begin{wraptable}{r}{0.6\textwidth}
\centering
% \vspace{-6mm}
% \scriptsize
\caption{Accuracy (\%) on Office-31 for UDA (ResNet-50). [$^{\dagger}$: average accuracy except $\text{D}\leftrightarrow \text{W}$.]  }
% \scalebox{0.85}{
\begin{adjustbox}{width=0.5\textwidth}
\begin{tabular}{l|cccccc|cc} 
\hline
Method & A$\to$D & A$\to$W & D$\to$A & D$\to$W & W$\to$A & W$\to$D  & Avg. {\cellcolor[rgb]{0.902,0.902,0.902}}& Avg.$^{\dagger}${\cellcolor[rgb]{0.902,0.902,0.902}}\\ 
\hline
ResNet-50   &78.3&70.4&57.3&93.4&61.5&98.1&76.5  {\cellcolor[rgb]{0.902,0.902,0.902}}& 66.9 {\cellcolor[rgb]{0.902,0.902,0.902}}\\ 
MinEnt~\cite{grandvalet2004semi}&90.7&89.4&67.1& 97.5& 65.0&\textbf{100.0}&85.0 {\cellcolor[rgb]{0.902,0.902,0.902}}&78.1 {\cellcolor[rgb]{0.902,0.902,0.902}} \\ 
MCC~\cite{jin2020minimum}&92.1&94.0&74.9& 98.5& 75.3 &\textbf{100.0}&89.1 {\cellcolor[rgb]{0.902,0.902,0.902}}&84.1 {\cellcolor[rgb]{0.902,0.902,0.902}} \\ 
BNM~\cite{cui2020towards}&92.2&94.0& 74.9& 98.5& 75.3 &\textbf{100.0}&89.2 {\cellcolor[rgb]{0.902,0.902,0.902}}&84.1  {\cellcolor[rgb]{0.902,0.902,0.902}}\\ 
PL &88.7&89.1&65.8 &98.1 &66.6 &99.6 &84.7  {\cellcolor[rgb]{0.902,0.902,0.902}}&77.6  {\cellcolor[rgb]{0.902,0.902,0.902}} \\ 
ATDOC-NC~\cite{liang2021domain}&95.2 &91.6 &74.6 &99.1 &74.7 &\textbf{100.0} & 89.2 {\cellcolor[rgb]{0.902,0.902,0.902}}& {\cellcolor[rgb]{0.902,0.902,0.902}}84.0  \\ 
ATDOC-NA~\cite{liang2021domain}&94.4 &94.3 &75.6 &98.9 &75.2 &99.6 &89.7  {\cellcolor[rgb]{0.902,0.902,0.902}}&84.9  {\cellcolor[rgb]{0.902,0.902,0.902}} \\ 
GeT &95.4 &95.4 & 76.6  & \textbf{99.1} &77.0 & \textbf{100.} &  90.6  {\cellcolor[rgb]{0.902,0.902,0.902}}& 86.0 {\cellcolor[rgb]{0.902,0.902,0.902}}  \\ 
\hline
CDAN+E~\cite{long2018conditional}&94.5 &94.2 &72.8 &98.6 &72.2 &\textbf{100.0} &88.7 {\cellcolor[rgb]{0.902,0.902,0.902}}&83.4  {\cellcolor[rgb]{0.902,0.902,0.902}} \\ 
+ MCC~\cite{jin2020minimum}&94.1 &94.7 & 75.4 & 99.0 &75.7  &\textbf{100.0} &89.8  {\cellcolor[rgb]{0.902,0.902,0.902}}&85.0  {\cellcolor[rgb]{0.902,0.902,0.902}}\\ 
+ BNM~\cite{cui2020towards}&94.9 &94.3  &75.8 &99.0 &75.9 &\textbf{100.0} &90.0  {\cellcolor[rgb]{0.902,0.902,0.902}}&85.2  {\cellcolor[rgb]{0.902,0.902,0.902}}\\ 
+ PL &91.5  &93.1 &72.5 &97.8 &72.7 &99.8 &87.9 {\cellcolor[rgb]{0.902,0.902,0.902}}&82.4 {\cellcolor[rgb]{0.902,0.902,0.902}}\\
+ ATDOC-NC~\cite{liang2021domain}&96.3 &93.6 &74.3 &\textbf{99.1} &75.4 &\textbf{100.0} &89.8. {\cellcolor[rgb]{0.902,0.902,0.902}}&84.9 {\cellcolor[rgb]{0.902,0.902,0.902}} \\ 
+ ATDOC-NA~\cite{liang2021domain}&95.4 &94.6 &77.5 &98.1 &77.0 &99.7 &90.4 {\cellcolor[rgb]{0.902,0.902,0.902}}&86.1 {\cellcolor[rgb]{0.902,0.902,0.902}} \\ 
+ GeT &\textbf{96.7} &\textbf{95.8} &\textbf{78.6}  &\textbf{99.1}  &\textbf{77.8} &\textbf{\textbf{100.}} &  \textbf{91.2}  {\cellcolor[rgb]{0.902,0.902,0.902}}& \textbf{87.2}  {\cellcolor[rgb]{0.902,0.902,0.902}} \\ 
\hline
MixMatch~\cite{berthelot2019mixmatch}&88.5 &84.6 &63.3 &96.1 & 65.0 & 99.6 &82.9  {\cellcolor[rgb]{0.902,0.902,0.902}}& 75.4  {\cellcolor[rgb]{0.902,0.902,0.902}}\\ 
w/ PL &89.0  &86.0  & 65.8 &96.2  & 65.6 & 99.6 &83.7  {\cellcolor[rgb]{0.902,0.902,0.902}}&76.6  {\cellcolor[rgb]{0.902,0.902,0.902}} \\ 
w/ ATDOC-NC~\cite{liang2021domain}&91.3 &86.4 &66.0 &97.4 &64.4 &99.4 &84.1  {\cellcolor[rgb]{0.902,0.902,0.902}}&77.0 {\cellcolor[rgb]{0.902,0.902,0.902}} \\ 
w/ ATDOC-NA~\cite{liang2021domain}&92.1  &91.0  &70.9  &98.6  &76.2  &99.6  &88.1  {\cellcolor[rgb]{0.902,0.902,0.902}}&82.6  {\cellcolor[rgb]{0.902,0.902,0.902}} \\ 
w/ GeT &93.1 &92.7 &71.8 &98.8 &77.0 &99.6 & 88.65  {\cellcolor[rgb]{0.902,0.902,0.902}}& 83.3 {\cellcolor[rgb]{0.902,0.902,0.902}}  \\ 
\hline
% SAFN+ENT~\cite{xu2019larger}&90.7/&90.1/&73.0/&98.6/&70.2/&99.8/&87.1/{\cellcolor[rgb]{0.902,0.902,0.902}}&81.0/{\cellcolor[rgb]{0.902,0.902,0.902}} \\ 
% CRST~\cite{zou2019confidence}&88.7/&89.4&72.6&98.9&70.9&\textbf{100.}&86.8{\cellcolor[rgb]{0.902,0.902,0.902}}&80.4 {\cellcolor[rgb]{0.902,0.902,0.902}}\\ 
SHOT~\cite{liang2020we}&94.0&90.1&74.7&98.4&74.3&99.9 &88.6{\cellcolor[rgb]{0.902,0.902,0.902}}&83.3{\cellcolor[rgb]{0.902,0.902,0.902}}\\ 
% CADA-P~\cite{kurmi2019attending}&95.6&\textbf{97.0}&71.5&\textbf{99.3}&73.1&\textbf{100.}&89.5{\cellcolor[rgb]{0.902,0.902,0.902}}&84.3{\cellcolor[rgb]{0.902,0.902,0.902}} \\ 
% ATM~\cite{li2020maximum}&96.4&95.7&74.1&\textbf{99.3}&73.5&\textbf{100.}&89.8{\cellcolor[rgb]{0.902,0.902,0.902}}&84.9 {\cellcolor[rgb]{0.902,0.902,0.902}}\\ 
SCDA~\cite{li2021semantic} &95.2 & 94.2 & 75.7 & 98.7 &76.2 & 99.8 & 90.0 {\cellcolor[rgb]{0.902,0.902,0.902}}& 85.3 {\cellcolor[rgb]{0.902,0.902,0.902}}\\
DALN~\cite{chen2022reusing} & 95.4 & 95.2 & 76.4 & 99.1 & 76.5 &\textbf{\textbf{100.}} & 90.4 {\cellcolor[rgb]{0.902,0.902,0.902}}& 85.9{\cellcolor[rgb]{0.902,0.902,0.902}}\\
\hline
\end{tabular}
% }
\end{adjustbox}
\label{tab: office-31}
% \vspace{-1mm}
\end{table}
% \vspace{-1mm}
% \end{wraptable}

\subsection{Comparison with Baselines}

% \vspace{-1mm}
\paragraph{Closed-set UDA.}
Tabs.~\ref{tab: office-home} and~\ref{tab: office-31} study the closed-set UDA setting using OfficeHome and Office-31 datasets under the standard setting.
% (see our supplementary for more UDA results on VisDA). 
We evaluate our method by combining it with three base models for comparisons: 1) Source-only model, trained with only labeled source data; 
2) CDAN+E, a UDA model with an additional domain alignment loss to train the model; 
3) MixMatch serving as the SSL base model.
We first study the regularization methods integrated with the Source-only model. BNM and MCC perform consistently better than the entropy regularization method, i.e. MinEnt, due to their design on the encouragement of prediction diversity generally ignored by entropy minimization. ATDOC that uses the target-oriented classifier significantly outperforms pseudo labeling (PL) and BNM, which verifies the importance of regularizing target predictions. %specially. 
As shown in Tabs.~\ref{tab: office-home} and~\ref{tab: office-31}, our GeT that uses the target structure similarity regularization
consistently achieves state-of-the-art performance. 
When combined with CDAN+E, all baselines show better results as the model is jointly optimized with an additional domain alignment loss.
Our GeT is able to further improve the performance 
% when further combined with the domain alignment methods CDAN+E, where  %accordingly.
and obtain the best average accuracy compared with other regularization methods. 
% It is further shown in Tab.~\ref{tab: office-31} and Tab.~\ref{tab: office-home} that GeT can achieve competitive results as some state-of-the-art UDA methods, e.g. ATM and SHOT, with no explicit feature alignment.
Under the SSL framework with MixMatch adopted as the base model,
our GeT also boosts the performance when it is adopted as a pseudo label generation module. 
% When the target domain is added with class distribution bias, the performance of all methods is inferior to their corresponding standard models suffering only from the data bias. 
%Our GeT designs the generative classifier which effectively alleviates the class bias by modeling the class prior and thus achieves more robust model performance.
% Nonetheless, our maximum entropy regularization on the label distribution in the target domain can still effectively alleviate the class distribution bias to achieve good performance.
It is further shown in Tab.~\ref{tab: office-home} and Tab.~\ref{tab: office-31} that GeT can achieve competitive results as some state-of-the-art UDA methods,~\eg SCDA and DALN, with no explicit feature alignment.

\begin{table*}[t!]
\caption{Classification accuracy (\%) on DomainNet-126 for SSDA (ResNet-34). } 
% \begin{center}
\centering
\begin{adjustbox}{width=1\textwidth}
% \begin{tabularx}{\textwidth}{l|XXXXXXXXXXXXXX|XX}
% \scriptsize
% \scalebox{0.3}{
\begin{tabular}{l|cccccccccccccc|cc}
% \begin{tabular*}{0.95\columnwidth}{@{\extracolsep{\stretch{1}}}*{1}{l|cccccccccccccc|cc}@{}}
% \begin{tabular}{l|cccccccccccccc|cc}
\hline
\multicolumn{1}{c|}{\multirow{2}{*}{Method}} & \multicolumn{2}{c|}{C$\to$S } & \multicolumn{2}{c|}{P$\to$C } & \multicolumn{2}{c|}{P$\to$R } & \multicolumn{2}{c|}{R$\to$C} & \multicolumn{2}{c|}{R$\to$P } & \multicolumn{2}{c|}{R$\to$S } & \multicolumn{2}{c|}{S$\to$P } & \multicolumn{2}{c}{Avg.{\cellcolor[rgb]{0.902,0.902,0.902}}}  \\
\cline{2-17}
\multicolumn{1}{c|}{}& \multicolumn{1}{c}{1-shot} & \multicolumn{1}{c|}{3-shot}  & \multicolumn{1}{c}{1-shot} & \multicolumn{1}{c|}{3-shot}  & \multicolumn{1}{c}{1-shot} & \multicolumn{1}{c|}{3-shot}    & \multicolumn{1}{c}{1-shot} & \multicolumn{1}{c|}{3-shot}   & \multicolumn{1}{c}{1-shot} & \multicolumn{1}{c|}{3-shot}   & \multicolumn{1}{c}{1-shot} & \multicolumn{1}{c|}{3-shot} & \multicolumn{1}{c}{1-shot} & \multicolumn{1}{c|}{3-shot}  &  \multicolumn{1}{c}{1-shot {\cellcolor[rgb]{0.902,0.902,0.902}}} & \multicolumn{1}{c}{3-shot {\cellcolor[rgb]{0.902,0.902,0.902}}}  \\
\hline
ResNet-34 &54.8 &57.9 &59.2 &63.0 &73.7 &75.6 &61.2 &63.9 & 64.5 &66.3 &52.0  &56.0  & 60.4 &62.2  &60.8  {\cellcolor[rgb]{0.902,0.902,0.902}}& 63.6  {\cellcolor[rgb]{0.902,0.902,0.902}}  \\
MinEnt~\cite{grandvalet2004semi} &56.3 &61.5 &67.7 &71.2 &76.0 &78.1 &66.1 &71.6 &68.9 &70.4 &60.0 & 63.5 &62.9 &66.0  & 65.4 {\cellcolor[rgb]{0.902,0.902,0.902}}& 68.9  {\cellcolor[rgb]{0.902,0.902,0.902}}\\
BNM~\cite{cui2020towards} &58.4 &62.6 &69.4 &72.7 &77.0 &79.5 &69.8 &73.7 &69.8 &71.2 &61.4  &65.1 &64.1  &67.6  &67.1  {\cellcolor[rgb]{0.902,0.902,0.902}}& 70.3  {\cellcolor[rgb]{0.902,0.902,0.902}}\\
MCC~\cite{jin2020minimum} &56.8 &60.5 &62.8 &66.5&75.3 &76.5 &65.5 &67.2 &66.9 &68.1 &57.6&59.8 &63.4 &65.0 &64.0  {\cellcolor[rgb]{0.902,0.902,0.902}}&66.2  {\cellcolor[rgb]{0.902,0.902,0.902}} \\
PL &62.5  &64.5 &67.6 1&70.7 &78.3 &79.3 &70.9 &72.9 &69.2 &70.7 &62.0 &64.8 &67.0  &68.6  &68.2  {\cellcolor[rgb]{0.902,0.902,0.902}}& 70.2 {\cellcolor[rgb]{0.902,0.902,0.902}} \\
ATDOC-NC~\cite{liang2021domain} &58.1 &62.2 &65.8 &70.2 &76.9 &78.7 &69.2 &72.3 &69.8 &70.6 &60.4  &65.0 &65.5  &68.1  &66.5  {\cellcolor[rgb]{0.902,0.902,0.902}}&69.6  {\cellcolor[rgb]{0.902,0.902,0.902}} \\
% ATDOC-NA~\cite{liang2021domain} &65.6 &66.7 &72.8 &74.2 &81.2 &81.2 &74.9 &76.9 &71.3 &72.5 &65.2  & 64.6 & 68.7 & 70.8 & 71.4 {\cellcolor[rgb]{0.902,0.902,0.902}}& 72.4  {\cellcolor[rgb]{0.902,0.902,0.902}}\\
GeT& \textbf{66.7}  &\textbf{67.8} &\textbf{73.9} &75.8 &\textbf{82.0}   & \textbf{82.8}   &  \textbf{76.1}  & \textbf{77.6}   & \textbf{72.5}  &\textbf{73.9} &  \textbf{66.8}  &67.1 & \textbf{69.8}   &  \textbf{73.6}   & \textbf{72.2}  {\cellcolor[rgb]{0.902,0.902,0.902}}& \textbf{73.9}  {\cellcolor[rgb]{0.902,0.902,0.902}}\\
\hline
MME~\cite{saito2019semi}&56.3&61.8&69.0&71.7&76.1&78.5&70.0&72.2&67.7&69.7&61.0&61.9&64.8&66.8&66.4{\cellcolor[rgb]{0.902,0.902,0.902}}&68.9{\cellcolor[rgb]{0.902,0.902,0.902}}\\
APE~\cite{kim2020attract}&56.7& 63.1&72.9 &76.7&76.6 &79.4 &70.4 &76.6 &70.8 &72.1 &63.0& 67.8 &64.5 &66.1& 67.6 {\cellcolor[rgb]{0.902,0.902,0.902}}& 71.7  {\cellcolor[rgb]{0.902,0.902,0.902}}\\
S$^3$D~\cite{yoon2022semi}& 60.8 &64.4 &73.4&75.1 &79.5 &80.3 &73.3 &75.9 &68.9 &72.1 &65.1 &66.7 &68.2 &70.0 &69.9{\cellcolor[rgb]{0.902,0.902,0.902}}& 72.1  {\cellcolor[rgb]{0.902,0.902,0.902}}\\
\hline
\end{tabular}
% }
\end{adjustbox}
% \end{tabularx}
% \end{center}
\label{tab: domainnet_ssda}
% \vspace{-3mm}
\end{table*}

\begin{table*}[t!]
\caption{Classification accuracy (\%) on Office-Home for PDA (ResNet-50).} 
% \begin{center}
\centering
\scriptsize
% \scalebox{0.6}{
\begin{adjustbox}{width=1\textwidth}
\begin{tabular}{l|cccccccccccc|c}
% \begin{tabular*}{0.95\columnwidth}{@{\extracolsep{\stretch{1}}}*{1}{l|ccccccccccccc}@{}}
\hline
Method & A$\to$C  & A$\to$P  & A$\to$R  & C$\to$A  & C$\to$P  & C$\to$R  & P$\to$A  & P$\to$C  & P$\to$R  & R$\to$A  & R$\to$C  & R$\to$P  &  Mean{\cellcolor[rgb]{0.902,0.902,0.902}}\\
\hline
ResNet-50 &43.5 &67.8 &78.9 &57.5&56.2 &62.2&58.1 &40.7 &74.9 &68.1 &46.1 &76.3 &60.9  {\cellcolor[rgb]{0.902,0.902,0.902}}\\
MinEnt~\cite{grandvalet2004semi}&45.7 &73.3 &81.6 &64.6 &66.2 &73.0 &66.0 &52.4 &78.7 &74.8 &56.7 &80.8 &67.8  {\cellcolor[rgb]{0.902,0.902,0.902}}\\
BNM~\cite{cui2020towards}&54.6 &77.2 &81.1 &64.9 &67.9 &72.8 &62.6 &55.7 &79.4 &70.5 &54.7 &77.6 &68.2  {\cellcolor[rgb]{0.902,0.902,0.902}}\\
MCC~\cite{jin2020minimum}&54.1 &75.3 &79.5 &63.9 &66.3 &71.8 &63.3 &55.1 5&78.0 &70.4 &55.7 &76.7 & 67.5 {\cellcolor[rgb]{0.902,0.902,0.902}}\\
PL &51.9 &70.7 &77.5 &61.7 &62.4 &67.8 &62.9 &54.1 &73.8 &70.4 &56.7 &75.0 &65.4 {\cellcolor[rgb]{0.902,0.902,0.902}}\\
ATDOC-NC~\cite{liang2021domain}&59.5 &80.3 &83.8 &71.8 &71.6 &79.7 &70.6 &59.4 &82.2 &78.4 &61.1 &81.5 &73.3 {\cellcolor[rgb]{0.902,0.902,0.902}}\\
ATDOC-NA~\cite{liang2021domain}&60.1 &76.9 &84.5 &72.8 &71.2 &80.9 &73.9 &61.8 &83.8 &77.3 &60.4 &80.4 &73.7  {\cellcolor[rgb]{0.902,0.902,0.902}}\\
GeT &\textbf{61.4} &\textbf{81.2} &\textbf{85.9}  &\textbf{74.0} &\textbf{74.1}  &\textbf{82.3} & \textbf{75.8}  & \textbf{63.9} & \textbf{85.3}  & \textbf{79.6}  & \textbf{63.7} & \textbf{84.6} & \textbf{75.8}  {\cellcolor[rgb]{0.902,0.902,0.902}}\\
\hline
SAFN~\cite{xu2019larger} & 58.9& 76.3 &81.4 &70.4&73.0& 77.8& 72.4&55.3 &80.4 &75.8&60.4& 79.9&  71.8{\cellcolor[rgb]{0.902,0.902,0.902}}\\
RTNet$_{adv}$~\cite{chen2020selective}&63.2&80.1 &80.7& 66.7& 69.3 &77.2 &71.6 &53.9&84.6 &77.4& 57.9& 85.5& 72.3{\cellcolor[rgb]{0.902,0.902,0.902}}\\
\hline
\end{tabular}
% }
% \end{center}
\end{adjustbox}
\label{tab: office-home-pda}
% \vspace{-3mm}
\end{table*}

% \vspace{-3mm}
\paragraph{Semi-supervised DA.} 
We follow the experiment setting in~\cite{saito2019semi} to evaluate SSDA on the DomainNet-126 dataset. As shown in Tab.~\ref{tab: domainnet_ssda}: 1) 1-shot represents one labeled instance is available for each class in the target domain, and 2) 3-shot means we have access to three target labels per class. From the reported results, pseudo-labeling and BNM achieve the same second best performance in the 3-shot setting while BNM performs better in the 1-shot setting.
% The baseline regularization methods exhibit unstable effects when further dealing with the bias from the target class distribution. 
The overall average accuracy of ATDOC-NC indicates that 
the performance of nearest centroid classifier relies heavily on the assumption of balanced target clusters to assign pseudo labels. 
We also include the prior state-of-the-art SSDA method, \eg S$^3$D, for comparison, which outperforms the SSL regularization approaches. 
By contrast, our GeT achieves the highest average accuracy among all the compared methods, which shows that our design on the target data structure learning indeed improves data discrimination.

% \vspace{-3mm}
\paragraph{Partial-set UDA.}
We follow the partial-set UDA setting in~\cite{liang2021domain} and evaluate performance on the OfficeHome dataset in Tab.~\ref{tab: office-home-pda} by selecting the first 25 classes as the label space for the unlabeled target data. 
PDA suffers from both the data bias and the label distribution shift, i.e., two domains have mismatched label space.
% ,  which makes the DA task more challenging.
% The class imbalance setting further adds class distribution bias in the target domain,  which makes the DA task more challenging.
% We further add class distribution bias into the target domain to make the DA task even more challenging.
ATDOC shows relative better results than other SSL regularization baselines as well as the prior state-of-the-art PDA methods, \ie RTNet$_{adv}$. 
MCC and BNM show comparable performance as MinEnT in the standard setting, but MinEnT achieves better results due to its superiority from the prediction diversity. 
Similarly, the structural similarity regularization in our GeT can penalize the over-confident predictions and shows effectiveness in improving performance.

\begin{table}[th]
% \vspace{1mm}
% \begin{wraptable}{r}{0.48\textwidth}
\caption{Classification accuracy (\%) on Office-Home for UDA and PDA under the imbalanced target distribution (ResNet-50). } 
\centering
% \scriptsize
\begin{adjustbox}{width=0.48\textwidth}
\begin{tabular}{l|ccccc|ccccc} 
\hline
Setting & \multicolumn{5}{c|}{UDA}  & \multicolumn{5}{c}{PDA} \\ 
\hline
\multicolumn{1}{l|}{Method} & A$\to$ C  & C$\to$P  & P$\to$R  & R$\to$A  & {\cellcolor[rgb]{0.902,0.902,0.902}}Avg. & A$\to$ C  & C$\to$P  & P$\to$R  & R$\to$A  & {\cellcolor[rgb]{0.902,0.902,0.902}}Avg.  \\ 
\hline
ResNet-50   &44.3 &62.4 &72.6 &64.3 & 59.5  {\cellcolor[rgb]{0.902,0.902,0.902}}  &49.1 &59.9 &76.1 &70.2 & 63.8 {\cellcolor[rgb]{0.902,0.902,0.902}}    \\
BNM~\cite{cui2020towards}  &55.9  &70.9&78.7  &70.4  & 67.5  {\cellcolor[rgb]{0.902,0.902,0.902}}  &53.8 & 63.7 &78.9  & 70.7 & 67.6  {\cellcolor[rgb]{0.902,0.902,0.902}}    \\ 
MCC~\cite{jin2020minimum}  &48.5  &66.8  &75.1 &67.6 & 63.1  {\cellcolor[rgb]{0.902,0.902,0.902}}  &52.4 & 61.1 & 75.7 & 70.1 & 65.6  {\cellcolor[rgb]{0.902,0.902,0.902}}    \\
PL &53.8  & 68.5 & 76.5 &69.2  & 65.4  {\cellcolor[rgb]{0.902,0.902,0.902}}  &49.9   &61.5  &72.8  &68.0  & 62.9  {\cellcolor[rgb]{0.902,0.902,0.902}}    \\ 
ATDOC-NC~\cite{liang2021domain}  &52.5 & 72.5 &78.6 &69.7 & 67.1 {\cellcolor[rgb]{0.902,0.902,0.902}}&55.6 & 71.8 &81.6 &73.9 & 70.9  {\cellcolor[rgb]{0.902,0.902,0.902}}    \\ 
% ATDOC-NA~\cite{liang2021domain}  &59.1 &46.6 &78.4 &75.9 & 65.0  {\cellcolor[rgb]{0.902,0.902,0.902}}  &55.9 &63.3  & 78.3 & 76.8 & 68.4  {\cellcolor[rgb]{0.902,0.902,0.902}}    \\
GeT& \textbf{56.1}  & \textbf{74.8} & \textbf{81.9}  & \textbf{71.6}  & \textbf{70.2}  {\cellcolor[rgb]{0.902,0.902,0.902}}  & \textbf{57.8}  & \textbf{73.8}  & \textbf{85.9}  & \textbf{77.2}  & \textbf{75.3}   {\cellcolor[rgb]{0.902,0.902,0.902}}    \\ 
\hline
\end{tabular}
\end{adjustbox}
% \end{center}
\label{tab: offhome-imbalance}
% \vspace{-1mm}
\end{table}
% \end{wraptable}

% \vspace{-3mm}
\paragraph{Imbalanced Target Distribution.} 
% Since the previous baselines are strictly under the assumption of the well-balanced class distribution, 
We further evaluate our method for the closed-set UDA and PDA under the imbalanced target label distribution scenario. As shown in Tab.~\ref{tab: offhome-imbalance}, when the target domain is added with class distribution bias, the performance of all methods is inferior to their corresponding standard models suffering only from the data bias,~\eg the performance of ATDOC deteriorates to be even more inferior than BNM. Our Get achieves the best results in all DA tasks and shows superior
resilience to severe label distribution shift. 

% \vspace{-3mm}
% \paragraph{Scare-labeled SSL.} 
% We further study a special SSL case where very few labels are given. We simply adopt the same 3-shot setting as SSDA in the target domain but remove the source domain data. We report results on the OfficeHome and DomainNet-126 datasets in Tab.~\ref{tab: offhome-domain-ssl}, and show the superior performance of our GeT. As the labels are quite limited, pseudo labeling only obtains accuracy that is roughly equivalent to the baseline. 

\begin{figure*}[t!]
% \vspace{-2mm}
\centering
\makebox[0.95\textwidth][c]{\includegraphics[scale=0.6]{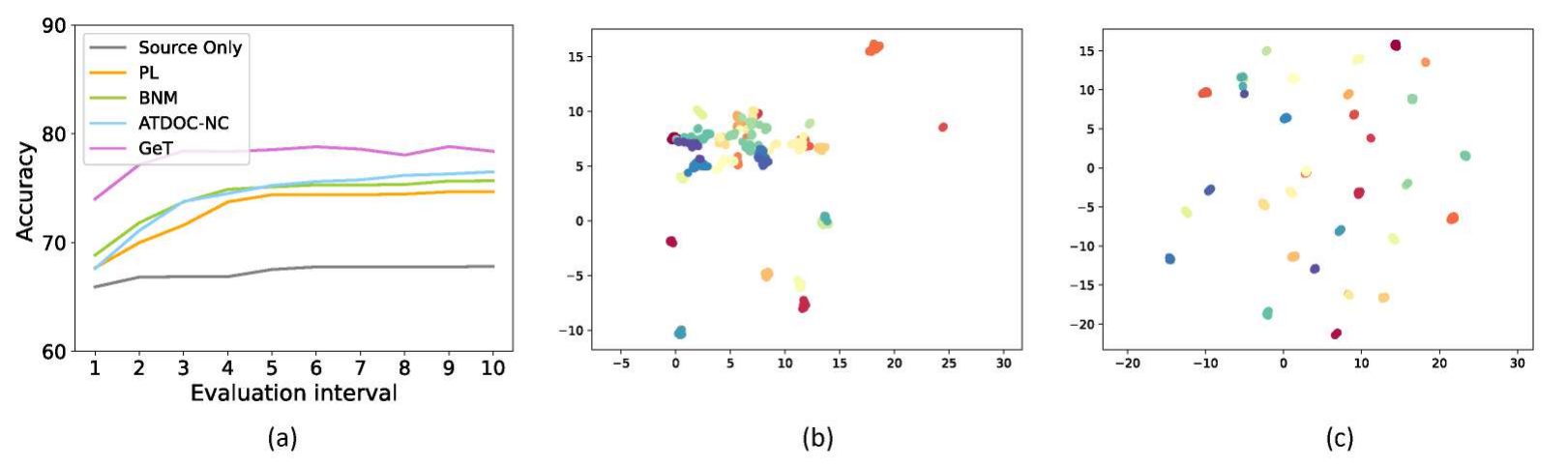}}
% \vspace{-3mm}
\caption{a) Comparison of convergence for the Ar $\to$ Pr task on Office-Home. Feature visualization with b) Source-only and c) GeT for the task A $\to$ W on Office-31. Note that different classes are denoted by different colors.
} 
% \vspace{-3mm}
\label{fig: visual}
\end{figure*}

\subsection{Model Analysis}

% \vspace{-3mm}
\paragraph{Ablation study.}
We conduct ablation study on Office-31 and VisDA-2017 for UDA in Tab.~\ref{tab: abalation} to examine the effect of each component on our GeT. The base model is ATDOC-NC where a target-oriented prototype classifier is used to generate pseudo labels. 1) \textit{Online update strategy for the probabilistic model} (i.e. w/o $\mathcal{L}_\text{KL}^{f, g}$). We present results by directly using the predictions $\hat{y}_{\mathbbm{M}}^t$ from the online updated generative classifier as pseudo labels.
% and removing the feature discrimination objectives $\mathcal{L}_{kl}^t$.  in the cross-entropy loss 
Compared to the pure target feature classifier (NC), our GeT improves $+1.1\%$ average accuracy on VisDA-2017 by modeling feature distributions with the generative classifier. It verifies the effectiveness of our online update strategy for the probabilistic model. 
2) \textit{Effect of feature structure regularization}. We examine the effect of each loss by removing each feature discrimination objective formulated by KL-divergence. We first evaluate $\mathcal{L}_{\text{KL}}^g$ in the label space (i.e. w/o $\mathcal{L}_{\text{KL}}^f$). It is shown the auxiliary distribution variable $Q_{g,f}$ can bring better performance than the oracle hard supervisions. The ensemble of feature-level regularization further improves $+1.7\%$ on VisDA-2017, thus showing the effectiveness of our %introduced 
learnable embedding prototypes for improving target feature discrimination. 3) \textit{Mixed soft labels}. We further analyze how pseudo labels generated from the maintained generative classifier improve upon the auxiliary label $Q_{g,f}$ with mixed supervisions (i.e. w/o $\hat{Y}_T$). We can see there is a performance drop (VisDA-2017: $83.4\%\to 82.0\%$) by removing the guidance from the online generative classifier. The mixed soft labels performs better than each separate supervision from $\hat{Y}_\mathbbm{M}$ and $Q_{g,f}$.

\begin{table}[t!]
% \begin{wraptable}{r}{0.5\textwidth}
% \vspace{-1mm}
\caption{Classification accuracy (\%) of GeT on Office-31 and VisDA-2017 under different variants. (ResNet-50)} 
\centering
\scriptsize
\begin{adjustbox}{width=0.45\textwidth}
\begin{tabular}{l|cccccc}
\hline
 & NC & GeT w/o $\mathcal{L}^{\{f, g\}}_\text{KL}$ &  GeT w/o $\mathcal{L}^{f}_\text{KL}$ & GeT w/o $\hat{Y}_T$ & GeT \\
\hline
Office-31& 84.0   & 85.1  &   84.5   & 85.6 & 86.0\\
VisDA-2017 & 80.3 & 81.4 & 80.6 & 82.0 & 83.4 \\
\hline
\end{tabular}
\end{adjustbox}
\label{tab: abalation}
% \vspace{-3mm}
\end{table}
% \vspace{-3mm}
% \end{wraptable}

% \vspace{-2mm}
\paragraph{Convergence comparison.}
We study the convergence of GeT by plotting test accuracy versus iteration number for Ar $\to$ Pr task on Office-Home in Fig.~\ref{fig: visual}. Comparing GeT with other baselines, GeT converges more quickly and the performance remains stable thereafter. %dose not show much fluctuation thereafter. 
This observation demonstrates that our GeT can provide more reliable pseudo labels in the early stage and performs better regularization on target data for discrimination. 

% \vspace{-2mm}
\paragraph{Visualization.}
We visualize the target features learned by Source Only and GeT for the task A $\to$ W on Office-31 in Fig.~\ref{fig: visual} with UMAP~\cite{umap} plot.
% Various colors indicate different categories.
It is obvious that our GeT provides better prototypes for target discrimination with clear boundaries among classes.

% \vspace{-2mm}
\paragraph{Sensitivity Analysis.}
\begin{figure}[t]
% \vspace{-5mm}
\centering
\noindent\includegraphics[scale=0.28]{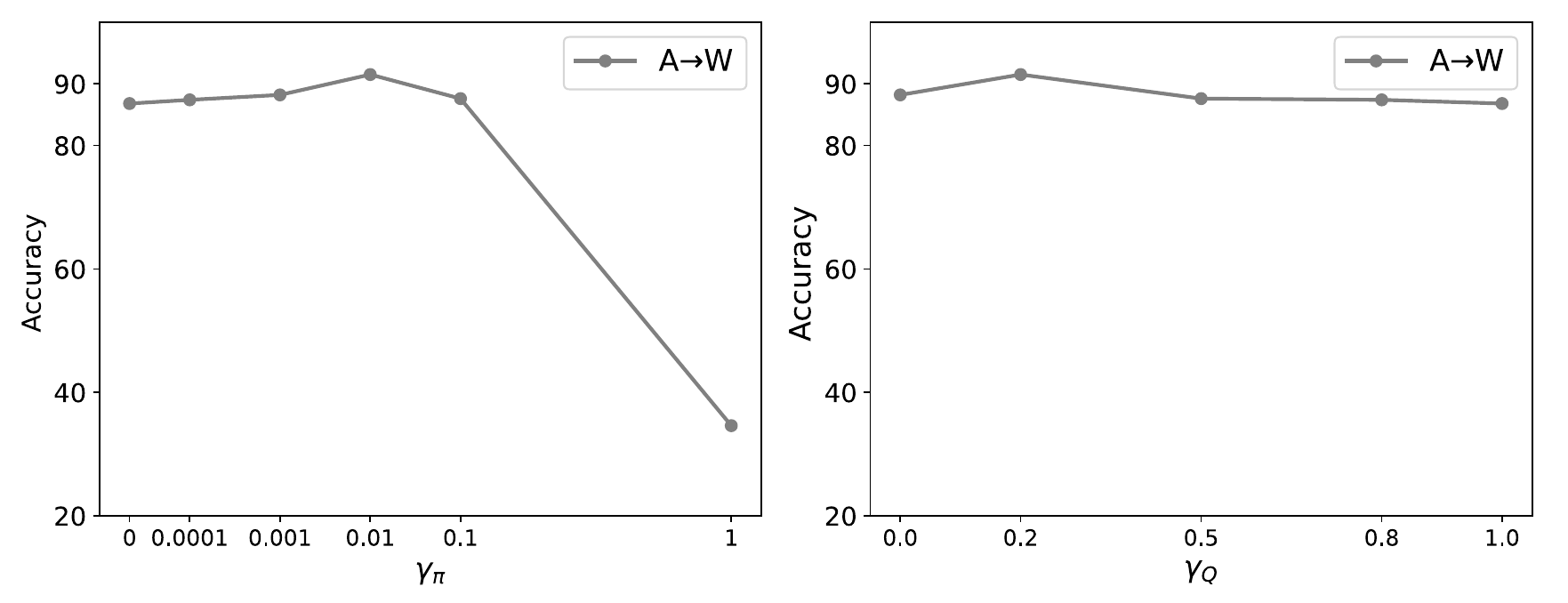}
% \vspace{-6mm}
\caption{Average accuracy of our GeT with a) varying $\gamma_\pi$ when $\gamma_Q=0.2$, and b) varying $\gamma_Q$ when $\gamma_\pi=0$ on A$\to$W in imbalanced Office-31.}
% \vspace{-2mm}
\label{fig: sensitivity}
\end{figure}

We also evaluate the sensitivity of our model to two hyper-parameters $\gamma_{\pi}$ and $\gamma_Q$, i.e. the memory decay for the class prior and the mixup coefficient for the soft labels in Fig.~\ref{fig: sensitivity}. Particularly, we choose $\gamma_\pi$ from [0,1] by setting $\gamma_Q$ to 0.2. we then vary the value of $\gamma_Q$ over the range \{0,0.2,0.5,0.8,1\} with $\gamma_\pi=0$.
% Fig.~\ref{fig: sensitivity} shows the accuracy of out GeT when varying the hyper-parameters. 
When $\gamma_Q$ is set to 0 or 1, it is equivalent to the single supervision from the model output or the target-oriented classifier. The mixed pseudo labels produce better results as they combine the source domain knowledge learned from the model and the target domain knowledge. It can be observed that the accuracy of our model is not sensitive to both hyper-parameters in a relative wide range.

\section{Conclusion}
\label{Conclusion}
In this paper, we propose a new target structure regularization approach for the DA tasks to deal with the source data bias and class distribution bias problems. We provide a new perspective of enhancing target data discrimination by formulating a learnable generative classifier, %where parameters are efficiently updated only based on the current observations.
where the parameters are updated efficiently online in mini-batches.
% To efficiently estimate parameters of the generative classifier, we propose an online update scheme by creating each component online only based on current observations. 
To further uncover the debiased target feature discrimination, we introduce the structure similarity regularization on the model predictions and the embeddings by an auxiliary distribution and a set of learnable embedding prototypes. Extensive experiments demonstrate that our GeT outperforms other regularization methods, and some DA models with explicit feature alignment on several DA tasks with large class distribution bias.

\paragraph{Acknowledgement.}
This research is supported by the National Research Foundation, Singapore under its AI Singapore Programme (AISG Award No: AISG2-RP-2021-024),
and the Tier 2 grant MOE-T2EP20120-0011 from the Singapore Ministry of Education.

{\small
\bibliographystyle{ieee_fullname}
\bibliography{egbib}
}

\end{document}